\documentclass[10pt,journal,compsoc]{IEEEtran}
\usepackage{bm}
\usepackage{amsfonts,amssymb}
\usepackage[mathscr]{euscript}
\usepackage{amsmath}
\DeclareMathAlphabet{\mathbit}{OT1}{cmr}{bx}{it}
\usepackage{graphicx}
\usepackage{float}
\usepackage{amssymb}
\usepackage{amsmath}
\DeclareMathOperator{\Tr}{Tr}

\usepackage{babel,blindtext}

\newcommand{\cov}{\mathrm{cov}}

\ifCLASSOPTIONcompsoc
  \usepackage[nocompress]{cite}
\else
  \usepackage{cite}
\fi

\ifCLASSINFOpdf
\else
\fi

\begin{document}
%
\title{Multi-output Gaussian Process Modulated Poisson Processes for Event Prediction}
%
%
%
%

\author{Salman~Jahani, Shiyu~Zhou*, Dharmaraj~Veeramani, and~Jeff~Schmidt
\IEEEcompsocitemizethanks{\IEEEcompsocthanksitem 
S. Jahani, S. Zhou (* Corresponding author) and D. Veeramani are with the Department of Industrial and Systems Engineering, University of Wisconsin-Madison, Madison,
WI 53706, USA.\protect\\
E-mail: jahani@wisc.edu; shiyuzhou@wisc.edu; raj.veeramani@wisc.edu
\IEEEcompsocthanksitem J. Schmidt is with the Raymond Corporation, Greene, NY 13778, USA\protect\\
E-mail: Jeff.Schmidt@raymondcorp.com
\IEEEcompsocthanksitem Under review at IEEE Transactions on Reliability.\\ 
\textcopyright 20xx IEEE. Personal use of this material is permitted. Permission from IEEE must be obtained for all other uses, in any current or future media, including reprinting/republishing this material for advertising or promotional purposes, creating new collective works, for resale or redistribution to servers or lists, or reuse of any copyrighted component of this work in other works.}
}


\IEEEtitleabstractindextext{%
\begin{abstract}
Prediction of events such as part replacement and failure events plays a critical role in reliability engineering.
Event stream data are commonly observed in manufacturing and teleservice systems. Designing predictive models for individual units based on such event streams is challenging and an under-explored problem. In this work, we propose a non-parametric prognostic framework for individualized event prediction based on the inhomogeneous Poisson processes with a multivariate Gaussian convolution process (MGCP) prior on the intensity functions. The MGCP prior on the intensity functions of the inhomogeneous Poisson processes maps data from similar historical units to the current unit under study which facilitates sharing of information and allows for analysis of flexible event patterns. To facilitate inference, we derive a variational inference scheme for learning and estimation of parameters in the resulting MGCP modulated Poisson process model. Experimental results are shown on both synthetic data as well as real-world data for fleet based event prediction.
\end{abstract}

}

\maketitle

\IEEEdisplaynontitleabstractindextext

%
\IEEEpeerreviewmaketitle

\IEEEraisesectionheading{\section{Introduction}\label{sec:introduction}}

\begin{figure}[ht]
\centering
  \includegraphics[scale=0.4]{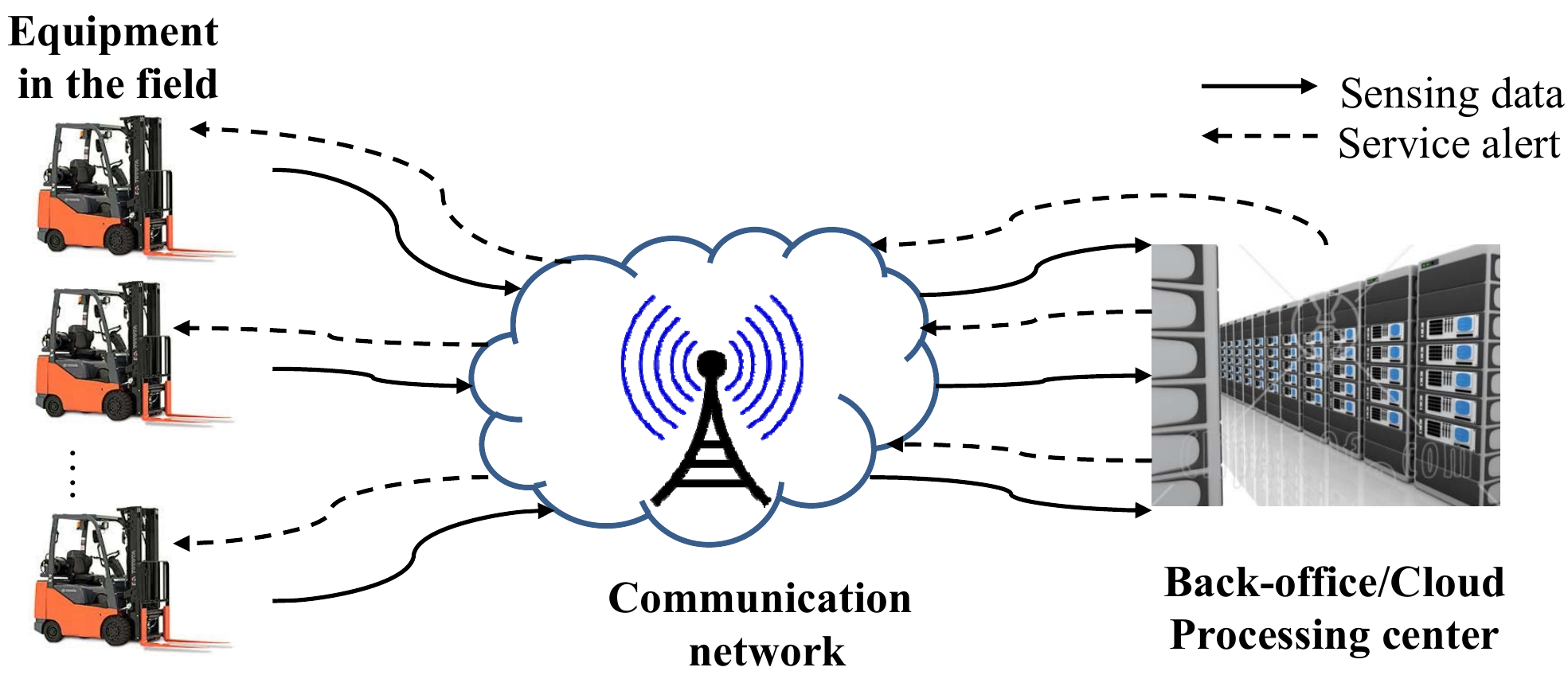}
  \caption{The structure of a teleservice system}
  \label{fig:tele}
\end{figure}

\IEEEPARstart{R}{ecent} advances in information and communication technology are playing a pivotal role in enabling what is referred to as Internet of Things (IoT). An example of IoT technology is teleservice systems. In a teleservice system, the data collected from a fleet of in-field units are transmitted through the communication network to the data processing center where the aggregated data are analyzed for condition monitoring and prognosis of in-field units. Figure \ref{fig:tele} illustrates such a teleservice system. 

Through the centralized data repository, the teleservice system has access to historical off-line records of events such as part replacements and failure events from all the units. The teleservice system also receives real-time event information from the in-field units. 
The availability of such a rich set of historical and real-time data in a teleservice system poses significant intellectual opportunities and challenges. On opportunities, since we have observations from potentially a very large number of similar units, we can compare their event patterns, share the information, and extract some common knowledge to enable accurate prediction at the individual level. On challenges, because the data are collected in the field and not in a controlled environment, the data contains significant variation and heterogeneity due to the large variations in working conditions for different units. This requires the developed analytics methods to be stochastic in nature to account for the variations. 

\begin{figure}[ht]
\centering
  \includegraphics[scale=0.45]{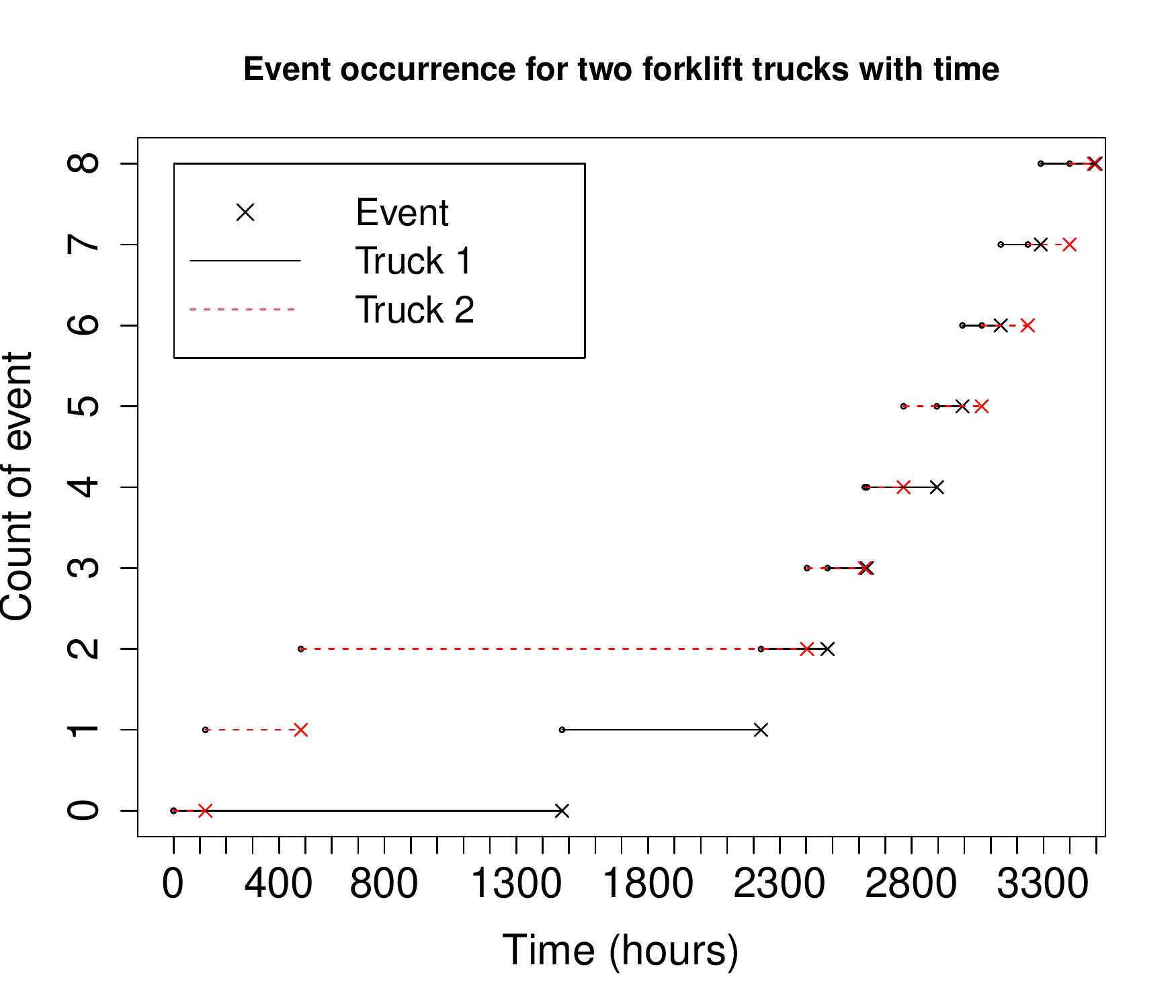}
  \caption{Illustration of event data from material handling forklifts.}
  \label{fig:casestudy}
\end{figure}

This work focuses on event prediction for individual units using the real-time event information collected from the unit under study as well as other units managed by the teleservice system. The event of interest occurs multiple times for each one of similar units during their lifetime. Figure \ref{fig:casestudy} illustrates typical event data of two forklifts collected in a teleservice system from a warehouse. An example of the event, here, could be replacement of a part due to its failure. In the figure, we can see that the event repeatedly occurs for each forklift. The pattern of occurrence for two forklifts bears some similarity but is distinct. One of the challenges in event prediction is how to extract useful information from data collected from other units to improve the prediction for the unit under study. This setting is known as \textit{multi-task learning}. The premise of this setting is that when multiple datasets from related outputs exist, their integrative analysis can be advantageous compared to learning multiple outputs independently. The goal of multi-task learning is to exploit commonalities between different units in order to improve the prediction and learning capabilities \cite{yuan2012visual,caruana1997multitask}. The key feature of multi-task learning is to provide a shared representation between training and testing outputs to allow inductive transfer of knowledge. In this paper, this inductive transfer of knowledge is achieved through specifying a valid semi definite covariance function that models dependencies of all data points \cite{conti2009gaussian}.

Events defined over a continuous domain arises in a variety of real-world applications including reliability analysis and event prediction for operational units/machines in connected manufacturing systems \cite{soleimani2017scalable}, disease prognosis in clinical trials \cite{jarrett2019dynamic} and events prediction using vital health signals from monitored patients at risk \cite{lian2015multitask}. One thread of work in such point process data focuses on learning event intensity rates by imposing smoothness on a latent rate function \cite{adams2009tractable,teh2011gaussian,lloyd2015variational}. Another consists of predicting future events as a direct function of past observations \cite{pillow2008spatio,gunawardana2011model}. Taking fleet based event prediction as a motivating example, we focus on the latter problem: given similar vehicles' history and the events history for the vehicle under-study up to time $t^\ast$, how many events will this vehicle have in $[t^\ast,t^\ast+L]$? Answering such a question provides a quantitative evaluation of the failure risk, which helps in making efficient maintenance plans and associated allocation of parts and resources.

Extensive research exists on event prediction, specially on failure prediction \cite{jardine2006review,si2011remaining}. The main avenue of research for event prediction using event data is focused on the data-driven statistical models. Data-driven statistical models typically estimate the probability of time-to-event distribution through parametric models (such as Weibull distribution) or non-parametric models \cite{meeker2014statistical,mcpherson2010reliability}. In this context, Cox PH regression has been widely used in clinical survival analysis and reliability engineering to investigate the effect of some covariates on the hazard rate/survival of a patient or a machine \cite{cox1972regression,klein2006survival}. Using the previously occurred events as covariates, Cox PH model has been used for event prediction using event stream data \cite{li2007failure,
yuan2011event}. The Cox PH model can incorporate several unit-specific factors as covariates in the regression. However, one limitation is that although the value of covariates can be unit specific, the model parameters are fixed and cannot reflect unit-to-unit variation. In other words, if the values of parameters for two units is the same, then the event prediction for these two units using the Cox PH model will always be the same. Also, the Cox PH regression model becomes inapplicable when good covariates are not available.

Another stream of research based on the event data uses frailty models as an extension of the conventional survival regression models, like Cox regression models, by incorporating a random effect term, typically called frailty term, to allow for unit-to-unit variation. Since the frailty term is random and follows a common distribution, the frailty model allows for unit-to-unit variation \cite{hougaard2012analysis,duchateau2007frailty, deep2019event}. However, the majority of research on frailty modeling is predominantly focused on investigating the significance of covariates and the frailty term in the fitted model rather than prediction for the individual units. In addition, frailty model is a parametric model that often needs relatively strong assumptions.

A few works have explored the event prediction problem in point processes by learning a functional mapping from history features to the current event intensity rate \cite{gunawardana2011model,pillow2008spatio,peng2018bayesian}. In Gunawardana et al. \cite{gunawardana2011model} the intensity function is constrained to be piecewise-constant, learned using decision trees and used for events' prediction. 
This setting is not appropriate for the events observation where the event intensity rate (or average incidence rate) varies smoothly over time. Moreover, this modeling approach does not take into account the variation between individual units. In our proposed approach, we will consider data where the intensity of the event generating process is assumed to vary smoothly over the domain. A popular model for such data is the inhomogeneous Poisson process with a Gaussian Process (GP) prior for the smoothly varying intensity function. This form of point processes are typically known as \textit{Cox processes} in literature \cite{kingmanc,moller1998log}. An example of such modeling approach is the Log Gaussian Cox Process (LGCP) where the log intensity function is driven by a GP prior \cite{moller1998log}. The flexibility of the LGCP comes at the cost of incredibly hard inference challenges due to its doubly stochastic nature and the notorious scalability issues of GP models. Another difficulty is that when available training data for each unit are scarce, then building such predictive models for event occurrence processes is difficult. This happens because industrial equipment nowadays tend to be generally very reliable and not subject to frequent failures. Therefore, we treat each individual event occurrence process as a task and follow a multi-task learning approach to share information from all tasks. This approach is in contrast to the general school of thought where a population level model is constructed \cite{meeker2014statistical,li2007failure,vrignat2015failure}.  Building a population model treats event prediction of different units' similarly. Such a population-level approach lacks the individualization capability where we need event predictions customized to an individual unit's history. The multi-task learning approach we propose here borrows information from the off-line historical event data and makes individualized predictions for a specific unit operating in the field.

Methods for learning GPs from multiple tasks have been proposed \cite{yu2005learning}, but they typically involve a shared global mean function and require inference at all observed data points across all the tasks. In the context of Cox processes, the inference of such multi-task models become even more challenging as it is doubly stochastic in nature and involves multiple correlated tasks \cite{diggle2013spatial,flaxman2015fast,leininger2017bayesian}. More details on double-stochasticity or double-interactability of Cox processes are given later in this study. One approximation method, variational Bayes, is often applied in such models leading to improvement in the scalablity. However, the inference in such predictive models lacks individualization capability. In this paper we propose a multi-task modeling approach enabling inference at individual level while sharing information from the historical offline data set.

The main objective of this study is to provide a framework for analysis of event occurrence probability of individual units under study. One challenge is that the available training data for each unit is typically sparse. Here, we propose a multivariate Gaussian convolution process (MGCP) modulated Poisson process model which facilities sharing of information from all units through a shared latent function.
The proposed framework borrows commonalities from different units and makes it possible to do inference and prediction at individual level. As mentioned before, a difficulty with building such a predictive model is that the inference is doubly-stochastic in nature and it scales poorly with the number of tasks and data points. Borrowing from the framework of the \textit{inducing variables} or \textit{pseudo inputs} in the GP literature \cite{snelson2006sparse,titsias2009variational}, we propose a variational inference framework to simultaneously estimate parameters in the resulting MGCP-Poisson Process (MGCP-PP) model. This facilitates the scalability and safeguards against model overfitting. Finally the advantageous feature of the proposed model is demonstrated through numerical studies and a case study with real-world data from forklift trucks' events. 

The remainder of this paper is organized as follows: In section 2, we provide an overview of the Cox processes. In section 3, we describe the problem formulation and inference scheme. In section 4 and section 5, we report the results of numerical studies and a real-world case study based on event data from a fleet of forklift trucks. Finally, our concluding remarks are given in section 6.

\section{Gaussian Process Modulated Poisson Process} \label{cox}

Assume data have been collected from $N$ units and let $I=\{1,2,\ldots ,N\}$ denote the set of all units. For unit $i \in I$, its associated data is $\bm{\mathcal{D}}_i=\{t_i^{(p)}\}_{p=1}^{P_i}$ where $t_i^{(p)}$ is the time that event $p$ occurred for unit $i$. Formally a Cox process –a particular type of inhomogeneous Poisson process– is defined via a stochastic intensity function $\lambda_i(t): \mathcal{X} \rightarrow \mathbb{R}^+$ for unit $i \in I$. For a domain $\mathcal{X}=\mathbb{R}$ where $\mathbb{R}$ is the real coordinate space, the number of points, $N(\mathcal{T})$, found in a subregion $\mathcal{T} \subset \mathcal{X}$ of unit $i$ is Poisson distributed with parameter $\lambda^{i}=\int_{\mathcal{T}}\lambda_i (t)dt$ and for disjoint subsets $\mathcal{T}_m$ of $\mathcal{X}$, the counts $N(\mathcal{T}_m)$ are independent. This independence is due to the completely independent nature of points in a Poisson process \cite{kingmanc}. 
If we restrict our consideration to some bounded region $\mathcal{T}$, the probability density of a set of $P_i$ observed points, $\bm{\mathcal{D}}_i$, conditioned on the rate function $\lambda_i(t)$ is 
\begin{equation}\label{eq:likelihood} 
p(\bm{\mathcal{D}}_i|{\lambda}_i)= \exp\left\{-\int_{\mathcal{T}} \lambda_i(u)du\right\} \prod_{p=1}^{P_i}\lambda_i (t_i^{(p)}).
\end{equation}
The likelihood of observed data across all $N$ units is $p(\bm{\mathcal{D}}|\bm{\lambda})=\prod_{i=1}^N p(\bm{\mathcal{D}}_i |{\lambda}_i)$ where $\bm{\mathcal{D}}=\{\bm{\mathcal{D}}_i\}_{i=1}^N$ and $\bm{\lambda}=\{\lambda_i\}_{i=1}^{N}$ is the collection of intensity functions for all units. 
Using Bayes' rule, the posterior distribution of the rate functions conditioned on the data, $p_d(\bm{\lambda}|\bm{\mathcal{D}})$, is:
\begin{equation}\label{eq:bayes} 
\frac{p_d(\bm{\lambda})\prod_{i=1}^N\exp\{-\int_{\mathcal{T}} \lambda_i(u)du\} \prod_{p=1}^{P_i}\lambda_i (t_i^{(p)})}{\int p_d(\bm{\lambda})\prod_{i=1}^N\exp\{-\int_{\mathcal{T}} \lambda_i(u)du\} \prod_{p=1}^{P_i}\lambda_i (t_i^{(p)})d\bm{\lambda}},
\end{equation}
which is often described as doubly-stochastic or doubly-intractable because of the nested integral in the denominator. Here we use the subscript $d$ to indicate the probability density function.

To overcome the challenges posed by the doubly-intractable integral, Adams et al. \cite{adams2009tractable} propose the Sigmoidal Gaussian Cox Process (SGCP). In the proposed SGCP model, a Gaussian process prior \cite{rasmussen2006gaussian} is used to construct an intensity function by passing a random function, $f\sim \mathcal{GP}$, through a sigmoid transformation and scaling it with a maximum intensity $\lambda^\ast$. The intensity function is therefore $\lambda (t)=\lambda ^\ast \sigma \left(f(t)\right)$, where $\sigma(.)$ is the logistic function
\begin{equation}\label{eq:logistic} 
\sigma(x)=\frac{1}{1+\exp(-x)}.
\end{equation}

While this model works well in practice in one dimension, in reality it scales poorly with the dimensionality of the domain and the number of datapoints. Moreover, it only considers the event data from the in-field unit and does not incorporate the off-line information from the historical units stored in the data repository. In order to tackle the issue of scalability, Lloyd et al. \cite{lloyd2015variational} propose to use a variational inference scheme. They assumed that the intensity is defined as $\lambda(t)=f^2(t)$ where $f\sim \mathcal{GP}$ is a GP distributed random function. This approach, also, falls short of considering the information that comes from the peer units in making inference and prediction, and only relies on the event data from the in-field unit. 

In this study we use a multi-task modeling approach to model the intensity functions of different units. This approach takes advantage of the multi-output GPs to share information between units from offline historical data and the online in-field unit via a shared latent function. A GP is formally defined as a collection of random variables, any finite number of which have consistent joint Gaussian distributions. For any input point $t \in \mathcal{X}\subset \mathbb{R}$, observations from a random dataset $\bm{f}(t)=\{f(t_1),f(t_2),...,f(t_p)\}^T$ are considered as single sample from some multivariate Gaussian distribution. Thus, the GP can be expressed as $f(t)\sim \mathcal{GP}(0,\Sigma(t,t'))$, where $\Sigma(t,t')$ is a positive definite covariance function. An alternative approach for constructing a Gaussian process is to convolve the GP random variables with an arbitrary kernel. Thus, $f(t)$ can be expressed as the convolution between a smoothing kernel $G(t)$ and a latent function $X(t)$ as follows:
\begin{equation}\label{eq:conv1} 
f(t)=\int_{\mathbb{R}}G(t-u)X(u)du.
\end{equation}
The resulting covariance function for $f(t)$ is then derived as 
\begin{equation}\label{eq:conv2} 
\cov_f(t,t')=\int_{\mathbb{R}}G(t-u)\int_{\mathbb{R}}G(t'-u')\kappa(u,u')du'du
\end{equation}
where $\cov[X(t),X(t')]=\kappa(t,t')$ is the covariance defining the latent function $X(t)$. We note that this construction is general in the sense that $X(t)$ can be any GP random variable \cite{alvarez2009sparse}. Therefore, the covariance matrix can be directly parametrized through parameters in the smoothing kernel. In this article we employ Convolution Process (CP) to build covariance functions that model dependencies within and across units. The basic idea is to build multiple GPs where all outputs depends on some common latent processes.  The proposed framework can provide each output with both shared and unique features and allows commonalities between different outputs to be automatically inferred. We introduce our multi-task modeling approach which takes advantage of the MGCPs in modeling the intensity of inhomogeneous Poisson processes in the next section. We also introduce a variational inference approach to make inference in the resulting MGCP modulated Poisson processes. The modeling framework introduced in this study makes inference and prediction for the individual in-field unit possible while tackling the sparsity in the observed event data.


\section{Construction of Multi-output Gaussian Process Modulated Poisson Process}

We construct our prior over the individual rate functions using GPs and assume that the resulting Cox process is driven by a latent log intensity function $\log \lambda_i := f_i$ with a GP prior:
\begin{equation}\label{eq:0}
f_i(t)\sim \mathcal{GP}\left(0,\Sigma_i(t,t')\right).
\end{equation}
To obtain an accurate predictive result, we need to capture relatedness among all $N$ units. Particularly, we use CP as mentioned in section \ref{cox} to model the latent log intensity functions $f_i (t)$ for each unit $i \in I$. We can consider a shared independent latent function $X(t)$ and $N$ different smoothing kernels $G_{i}(t):i=1,...,N$. The latent function is assumed a GP with covariance $\cov[X(t),X(t')]=\kappa(t,t')$. We set the kernels as 
\begin{equation}\label{eq:ker} 
\begin{gathered}
G_{i}(t)=\frac{a_i\pi^{-\frac{1}{4}}}{\sqrt{|\xi_i|}}\exp(-\frac{1}{2}\frac{t^2}{\xi_i^2}):=\alpha_{i}\mathcal{N}(t;0,\xi_{i}^2),
\end{gathered}
\end{equation}
to be scaled Gaussian kernels where $\mathcal{N}(t;0,\xi_{i}^2)$ is the density function of a zero mean normal distribution with variance $\xi_{i}^2$. We also consider $\kappa(t,t')$ to be the squared exponential covariance function \cite{alvarez2009sparse} as follows:
\begin{equation}\label{eq:1} 
\begin{split}
\kappa(t,t')&=\exp\left[-\frac{1}{2}\frac{(t-t')^2}{\lambda^2}\right]\\ 
&=\sqrt{2\pi\lambda^2}\mathcal{N}(d;0,\lambda^2):=C\mathcal{N}(d;0,\lambda^2),
\end{split}
\end{equation}

The GP $f_i(t)$ is then constructed by convolving the shared latent function with the smoothing kernel as follows:

\begin{equation}\label{eq:2}
\begin{gathered}
f_i(t)=\int_{\mathbb{R}}G_{i}(t-u)X(u)du.
\end{gathered}
\end{equation}

\begin{figure}
\centering
  \includegraphics[scale=0.09]{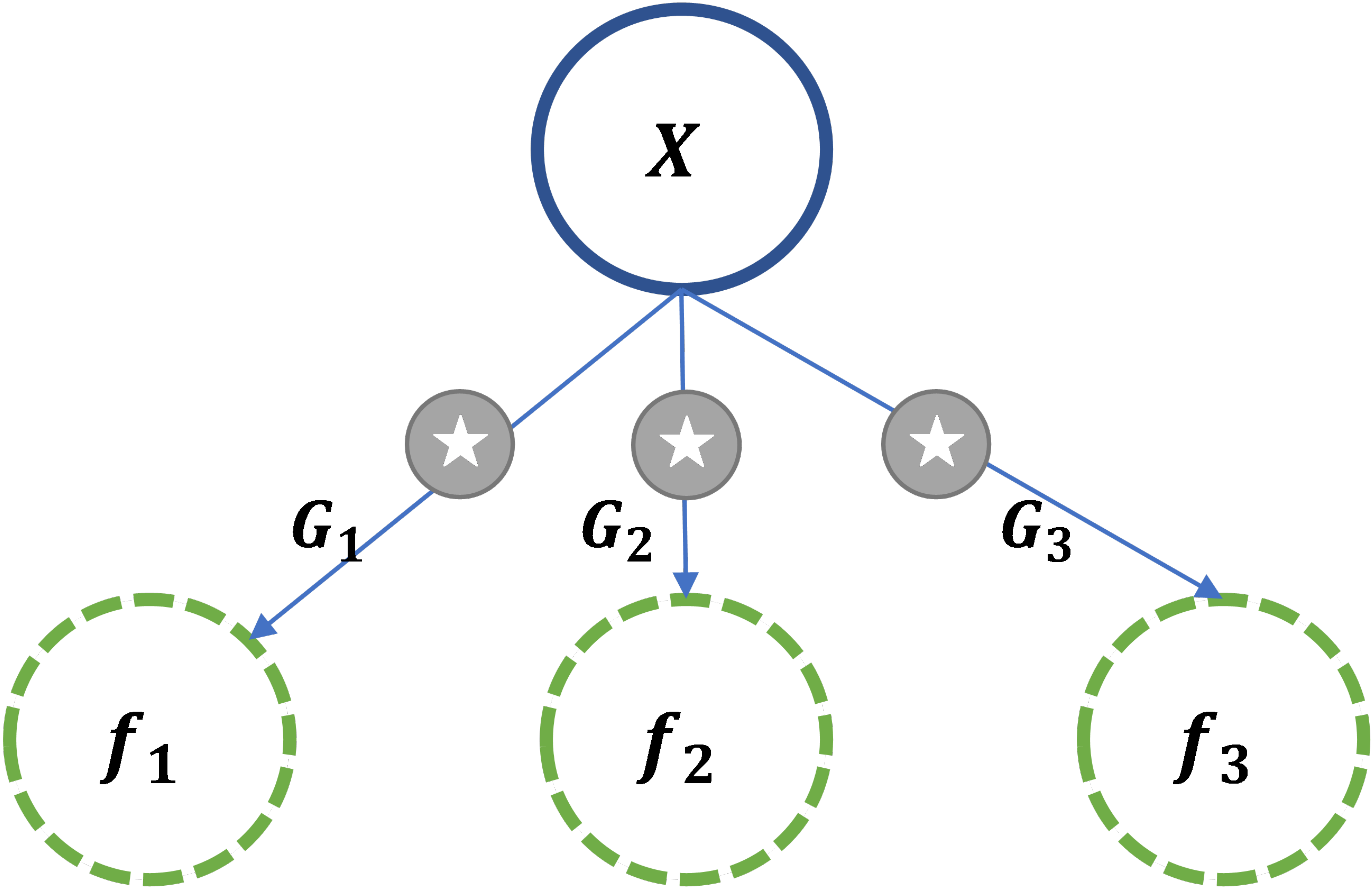}
  \caption{A convolution process with one latent functions}
  \label{fig:fig1}
\end{figure}

This is the underlying principle of MGCP, where the latent functions $X(t)$ is shared across different units through the corresponding kernel $G_{i}(t)$. Since the model in Eq. \eqref{eq:2} shares the latent function, a GP, across multiple units and since convolution is a linear operator, all outputs can be expressed as a jointly distributed GP. Figure \ref{fig:fig1} shows an illustration of such a convolution structure.
As shown in figure \ref{fig:fig1}, the key feature is that information is shared through parameters encoded in the kernels $G_{i}(t)$. Outputs then possess both unique and shared features; thus, accounting for heterogeneity in the intensity functions.

Based on equation \eqref{eq:2}, the covariance function between $f_i$ and $f_j$ and the covariance function between $f_i$ and $X$, can be calculated as follows:

\begin{equation}\label{eq:3}
\begin{split}
\cov_{f_i,f_j}(t,t')&=\int_{\mathbb{R}}G_{i}(t-u)\int_{\mathbb{R}}G_{i}(t'-u')\kappa(u,u')du'du\\
&=\alpha_{i}\alpha_{j}\sqrt{\frac{\lambda^2}{\eta_{i,j}^2}}\exp(-\frac{1}{2}\frac{(t-t')^2}{\eta^2_{i,j}}),\\
\cov_{f_i,X}(t,u)&=\int_{\mathbb{R}}G_{i}(t-u')\kappa(u,u')du'\\
&=\alpha_{i}\sqrt{\frac{\lambda^2}{\eta_{i}^2}}\exp(-\frac{1}{2}\frac{(t-u)^2}{\eta^2_{i}}),
\end{split}
\end{equation}
where $\eta^2_{i,j}=\xi^2_{i}+\xi^2_{j}+\lambda^2$ and $\eta^2_{i}=\xi^2_{i}+\lambda^2$. Now denote the underlying latent log intensity rates at the input data points as $\bm{f}=\{\bm{f}_1^T,...,\bm{f}_N^T\}^T$, where $\bm{f}_i=\{f_i(t_{i}^{(1)}),...,f_i(t_{i}^{(p_i)})\}^T$. The density function of $\bm{f}$ can be obtained as $p_d(\bm{f})=\mathcal{N}(\bm{f};\bm{0},\bm{K_{f,f}})$, where $\bm{K_{f,f}}$ sized $(\sum_{i=1}^Np_i)\times (\sum_{i=1}^Np_i)$ is the covariance function. 

Exact inference in the proposed model entails optimizing the \textit{model evidence} $p(\bm{\mathcal{D}})=\mathbb{E}_{p(\bm{f})}\left[p(\bm{\mathcal{D}}|\bm{\lambda}=\exp(\bm{f}))\right]$ for which the marginal log-likelihood can be obtained as follows:

\begin{equation}\label{evidence}
\log p(\bm{\mathcal{D}})=\log \int p\left(\bm{\mathcal{D}}|\bm{\lambda}=\exp(\bm{f})\right)p_d(\bm{f})d\bm{f},
\end{equation}
where as noted before $p_d(\bm{f})=\mathcal{N}(\bm{f};\bm{0},\bm{K_{f,f}})$. The likelihood of $\bm{f}$ in Eq. \eqref{evidence} involves inversion of the large matrix $\bm{K_{f,f}}$ which has a limiting cubic complexity $O\left(\left(\sum_{i=1}^Np_i\right)^3\right)$ and is in general intractable. Moreover, as mentioned in section \ref{cox}, we see that the log-likelihood is doubly-stochastic as it also involves an integration over the latent log intensity functions (see Eq. \eqref{eq:likelihood} and Eq. \eqref{evidence}). This, in turn, makes the exact inference more challenging. To alleviate the computation burden of matrix inversion, low-rank Gaussian process functions can be constructed by \textit{augmenting} the Gaussian process with a small number of $M$ inducing points or pseudo-inputs from the shared latent function. In next subsection, we introduce a variational inference framework based on the inducing points which tackles the double-stocasticity of Eq. \eqref{evidence} by obtaining a lower bound on the model evidence.

\subsection{Variational Inference} \label{variational}

We denote the inducing points by $\bm{\mathcal{Z}}=\{z_i\}_{i=1}^M$ and the value of shared latent function at the inducing points by $\bm{X}=[X(z_1),...,X(z_M)]^T$. Since the latent function is GP, any sample $\bm{X}$ follows a multivariate Gaussian distribution. Therefore, the probability distribution of $\bm{X}$ can be expressed as $p_d(\bm{X}|\bm{\mathcal{Z}})=\mathcal{N}(\bm{X};\bm{0},\bm{K_{X,X}})$, where $\bm{K_{X,X}}$ is constructed by the covariance function in equation \eqref{eq:1}. We now can sample from the conditional prior $p(X(u)|\bm{X},\bm{\mathcal{Z}})$. In equation \eqref{eq:2} where we construct latent intensity function $f_i(t)$, $X(u)$ can be well approximated by the expectation $\mathbb{E}(X(u)|\bm{X},\bm{\mathcal{Z}})$ as long as the latent function is smooth \cite{alvarez2009sparse}. By multivariate Gaussian identities, the probability distribution of $\bm{f}$ conditional on $\bm{X}$, $\bm{\mathcal{Z}}$ is:

\begin{equation}\label{eq:4}
\begin{split}
p_d(\bm{f}|\bm{X, \mathcal{Z}})=&\mathcal{N}(\bm{f};\bm{K_{f,X}}\bm{K_{X,X}^{-1}}\bm{X},\\
&\bm{K_{f,f}}-\bm{K_{f,X}}\bm{K_{X,X}^{-1}}\bm{K_{X,f}}),
\end{split}
\end{equation}
where $\bm{K_{X,X}}$ is the covariance matrix between the inducing variables and $\bm{K_{f,X}}$ is the covariance matrix between the latent log intensity values and the inducing variables. Therefore, $p_d(\bm{f})$ can be approximated by $p_d(\bm{f}|\bm{\mathcal{Z}})$, which is given as:
\begin{equation}\label{eq:5}
p_d(\bm{f}|\bm{\mathcal{Z}})=\int p_d(\bm{f}|\bm{X},\bm{\mathcal{Z}})p_d(\bm{X}|\bm{\mathcal{Z}})d\bm{X}.
\end{equation}
By equation \eqref{eq:5}, the marginal log-likelihood function can be approximated as follows:
\begin{equation}\label{eq:6}
\begin{split}
&\log p(\bm{\mathcal{D}})=\log \int p\left(\bm{\mathcal{D}}|\bm{\lambda}=\exp(\bm{f})\right)p_d(\bm{f})d\bm{f} \\
&\approx \log \int\int p\left(\bm{\mathcal{D}}|\bm{\lambda}=\exp(\bm{f})\right)p_d(\bm{f}|\bm{X, \mathcal{Z}})p_d(\bm{X|\mathcal{Z}})d\bm{X}d\bm{f}
\end{split}
\end{equation}

We next continue by integrating out the inducing variables $\bm{X}$, using a variational distribution $q_d(\bm{X})=\mathcal{N}(\bm{X};\bm{m},\bm{S})$ over the inducing points. We then multiply and divide the joint by $q_d(\bm{X})$ and lower bound using Jensen's inequality to obtain a lower bound on the model evidence:

\begin{equation}\label{eq:7}
\begin{split}
\log p(\bm{\mathcal{D}})&=\log \left[\int\int p(\bm{\mathcal{D}}|\bm{f})p_d(\bm{f}|\bm{X})p_d(\bm{X})\frac{q_d(\bm{X})}{q_d(\bm{X})}d\bm{X}d\bm{f} \right]\\
&\geq \int\int p_d(\bm{f}|\bm{X})q_d(\bm{X})d\bm{X}\log (p(\bm{\mathcal{D}}|\bm{f})) d\bm{f}\\
&+\int\int p_d(\bm{f}|\bm{X})q_d(\bm{X})d\bm{f}\log (\frac{p_d(\bm{X})}{q_d(\bm{X})}) d\bm{X}\\
&=\mathbb{E}_{q_d(\bm{f})}\left[\log p(\bm{\mathcal{D}}|\bm{f}) \right]-KL(q_d(\bm{X})\parallel p_d(\bm{X}))\triangleq\mathcal{L}
\end{split}
\end{equation}
Since $p_d(\bm{f}|\bm{X})$ is conjugate to $q_d(\bm{X})$, we can write down in closed form the resulting integral:

\begin{equation}\label{eq:8}
\begin{split}
q_d(\bm{f})&=\int p_d(\bm{f}|\bm{X})q_d(\bm{X})d\bm{X}:=\mathcal{N}(\bm{f};\bm{\mu},\bm{\Sigma})\\
\bm{\mu}&=\bm{K_{f,X}}\bm{K_{X,X}}^{-1}\bm{m}\\
\bm{\Sigma}&=\bm{K_{f,f}}-\bm{K_{f,X}}\bm{K_{X,X}}^{-1}(\bm{I}-\bm{S}\bm{K_{X,X}}^{-1})\bm{K_{X,f}}.
\end{split}
\end{equation}
Here, $KL(q_d(\bm{X})\parallel p_d(\bm{X}))$ is simply the KL-divergence between two Gaussians:
\begin{equation}\label{eq:9}
\begin{split}
KL(q_d(\bm{X})\parallel p_d(\bm{X}))=&\frac{1}{2}\left[ \Tr(\bm{K_{X,X}}^{-1}\bm{S})-\log \frac{|\bm{K_{X,X}}|}{|\bm{S}|}\right. \\
&\left.-M+(\vec{0}-\bm{m})^T\bm{K_{X,X}}^{-1}(\vec{0}-\bm{m}) \right],
\end{split}
\end{equation}
where $\Tr(.)$ is a trace operator. We now take expectation of data log-likelihood under $q_d(\bm{f})$:
\begin{equation}\label{eq:10}
\begin{split}
\mathcal{L}&=\mathbb{E}_{q_d(\bm{f})}\left[\log p(\bm{\mathcal{D}}|\bm{f}) \right]-KL(q_d(\bm{X})\parallel p_d(\bm{X}))\\
&=\mathbb{E}_{q_d(\bm{f})}\left[-\sum_{i=1}^N\int_{\mathcal{T}} \exp (f_i(u))du+\sum_{i=1}^N\sum_{p=1}^{P_i} f_i(t_i^{(p)})\right]\\
&-KL(q_d(\bm{X})\parallel p_d(\bm{X}))
\end{split}
\end{equation}
The first term of integration in \eqref{eq:10} can be analytically calculated as follows:
\begin{equation}\label{eq:11}
\begin{split}
\mathbb{E}_{q_d(\bm{f})}&\left[\sum_{i=1}^N\int_{\mathcal{T}} \exp (f_i(u))d u\right]\\
&=\sum_{i=1}^N\int\int q_d(\bm{f})\exp(f_i(u))d\bm{f}du\\
&=\sum_{i=1}^N\int \exp(\mu_i(u)+\frac{1}{2}\sigma^2_i(u) )du
\end{split}
\end{equation}
where $\mu_i(u):=\bm{K}_{f_i(u),\bm{X}}\bm{K}_{\bm{X,X}}^{-1}\bm{m}$ and $\sigma^2_i(u):=\bm{K}_{f_i(u),f_i(u)}-\bm{K}_{f_i(u),\bm{X}}\bm{K}_{\bm{X},\bm{X}}^{-1}(\bm{I}-\bm{S}\bm{K}_{\bm{X},\bm{X}}^{-1})$ $\bm{K}_{\bm{X},f_i(u)}$. The second term in \eqref{eq:10} can be calculated as follows:
\begin{equation}\label{eq:12}
\begin{split}
\mathbb{E}_{q(\bm{f})}\left[\sum_{i=1}^N\sum_{p=1}^{P_i} f_i(t_i^{(p)})\right]=\sum_{i=1}^N\sum_{p=1}^{P_i} \mu_i(t_i^{(p)})
\end{split}
\end{equation}
where $\mu_i(t_i^{(p)})=\bm{K}_{t_i^{(p)},\bm{X}}\bm{K}_{\bm{X},\bm{X}}^{-1}\bm{m}$. To perform inference we find the variational parameters $\bm{m}^\ast$, $\bm{S}^\ast$ and the model parameters $\bm{\theta}^{\ast}=(\{\lambda,\xi_{i},\alpha_{i}\}_{i=1}^{N})$ that maximize the $\mathcal{L}$. To optimize these simultaneously, we construct an augmented vector $\bm{\Theta}=\left[\bm{\theta},\bm{m}^T,vech(\bm{L})^T\right]$ where $vech(\bm{L})$ is the vectorization of the lower triangular elements of $\bm{L}$, such that $\bm{S}=\bm{L}\bm{L}^T$. The $vech(.)$ operator is a linear transformation which converts a matrix into a column vector.

\subsection{Predictive Distribution}

\begin{figure*}[!t]
\centering
  \includegraphics[scale=0.1]{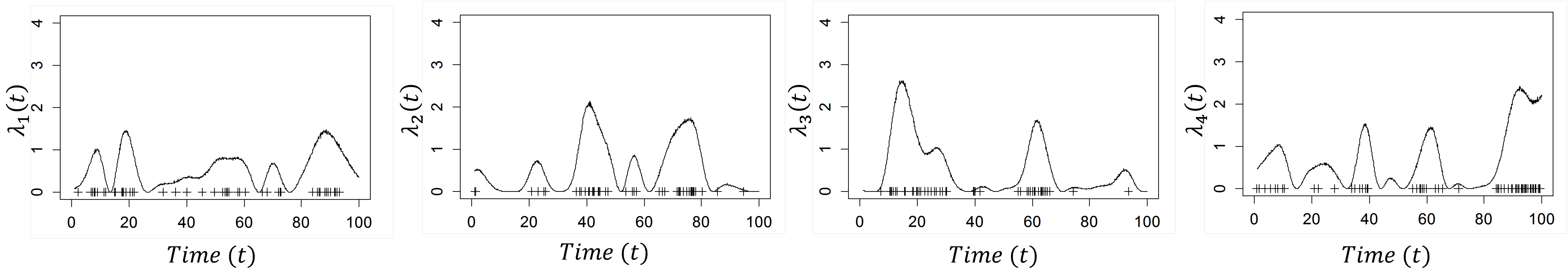}
  \caption{A sample of intensity rates generated from MGCP and the sigmoid link function.}
  \label{fig:fig2}
\end{figure*}

In this section, we derive the predictive distribution for the test unit $N$ based on the optimized $\bm{\Theta}^\ast$. Our training data (denoted as $\bm{\mathcal{D}}$) includes the observations from the offline units $i=1, 2, ..., N-1$ as well as the partial observations from the online test unit $N$.

Suppose observations from the test unit $N$ have been collected up to time $t^\ast$. We can next derive the predictive distribution for any new input time $T$ of the test unit $N$. In order to form the predictive distribution we assume our optimised variational distribution $q_d^\ast(\bm{X})=\mathcal{N}(\bm{X};\bm{m}^\ast,\bm{S}^\ast)$ approximates the posterior $p_d(\bm{X}|\bm{\mathcal{D}})$. Similar to Equation \eqref{eq:8}, we next compute $q_d^\ast(\bm{f})\approx p_d(\bm{f}|\bm{\mathcal{D}})$. We can now derive a lower bound of the (approximate) predictive log-likelihood for unit $N$ in any new input time $T$:
\begin{equation}\label{eq:13}
\begin{split}
\log p(T|\bm{\mathcal{D}},\bm{\Theta}^\ast)&=\log \mathbb{E}_{p_d(\bm{f}|\bm{\mathcal{D}})}[p(T|\bm{f})]\\
&\approx \log\mathbb{E}_{q_d^\ast(\bm{f})}[p(T|\bm{f})]\\
&\geq \mathbb{E}_{q_d^\ast(\bm{f})}[\log p(T|\bm{f})]\triangleq \mathcal{L}_p
\end{split}
\end{equation}
The derivation of $\mathcal{L}_p$ follows Equations \eqref{eq:10}-\eqref{eq:12}. The resulting bound is similar to $\mathcal{L}$ except that $\bm{m}$, $\bm{S}$ are replaced with $\bm{m}^\ast$ and $\bm{S}^\ast$, and there is no KL-divergence term. All the kernel matrices are computed using $\bm{\Theta}^\ast$. We use this bound to give results from approximate predictive likelihood when comparing against other approaches.

We can now answer the question posed at the beginning of this study using the derived predictive log-likelihood. The question involves estimating the distribution of event occurrence in $[t^\ast,t^\ast+L]$ for the test unit $N$. This event occurrence probability distribution for unit $N$ depends on the predicted latent log intensity function $\log \lambda_N := f_N(u), u\in[t^\ast,t^\ast+L]$. Given the predicted intensity rate $\lambda^{N^\ast}=\int_{t^\ast}^{t^\ast+L}\lambda_{N}(t)dt$, the event occurrence has Poisson distribution. Given the estimated parameters, we are interested in:
\begin{equation}\label{eq:14}
\begin{split}
p_d(N(t^\ast)=y)=\frac{e^{-\lambda^{N^\ast}}{\lambda^{N^\ast}}^y}{y!}, \  y=0,1,2,...,
\end{split}
\end{equation}
where $y$ is the number of events. Based on \eqref{eq:14}, the accurate probability of event occurrence depends on the extrapolation of the intensity rate within $L$ for the testing unit $N$. In the MGCP, the predictive distribution for any new input point $T^\ast$ is given by:

\begin{equation}\label{eq:15}
\begin{split}
p_d(f_N(T^\ast)|\bm{\mathcal{D}})&=\int p_d(f_N(T^\ast)|\bm{X})p_d(\bm{X}|\bm{\mathcal{D}})d\bm{X}\\
&\approx \int p_d(f_N(T^\ast)|\bm{X})q_d(\bm{X})d\bm{X}\\
&=\mathcal{N}\left(f_N(T^\ast);\bm{K}_{f_N(T^\ast),\bm{X}}\bm{K_{X,X}}^{-1}\bm{m}^\ast,\right.\\ 
&\bm{K}_{f_N(T^\ast),f_N(T^\ast)}-\bm{K}_{f_N(T^\ast),\bm{X}}\bm{K}_{\bm{X},\bm{X}}^{-1}\times\\
&\left.(\bm{I}-\bm{S}^\ast\bm{K}_{\bm{X},\bm{X}}^{-1})\bm{K}_{\bm{X},f_N(T^\ast)}\right)
\end{split}
\end{equation}
where we assumed our optimized variational distribution $q^\ast(\bm{X})=\mathcal{N}(\bm{X};\bm{m}^\ast,\bm{S}^\ast)$ approximates the posterior $p_d(\bm{X}|\bm{D})$. We used $\bm{K}_{f_N(T^\ast),f_N(T^\ast)}$ as a notation when the covariance matrix is evaluated at $T^\ast$. Consequently, the predictions at the time point $T^\ast$ for unit $N$ is $\hat{f}_N(T^\ast)=\bm{K}_{f_N(T^\ast),\bm{X}}\bm{K_{X,X}}^{-1}\bm{m}^\ast$.


\section{Experiments}

In this section, the performance of our proposed methodology, denoted as MGCP-PP is investigated. We benchmark the prediction performance of our proposed framework using both synthetic and real-world data. Specifically, we benchmark the  performance against Variational
Bayes for Point Processes (VBPP) approach of Lloyd et al.  \cite{lloyd2015variational} and the Sigmoidal Gaussian
Cox Process (SGCP) of Adams et al. \cite{adams2009tractable} which are based on considering a univariate Gaussian process prior for the intensity rate. Unlike our proposed approach, the methods discussed in \cite{lloyd2015variational} and \cite{adams2009tractable} do not consider the cross-correlation that exists between different units. Regarding our MGCP-PP model we set the number of pseudo-inputs to $M=10$. Throughout this section we consider $N=10$ units and it is assumed that the observations are made in $t\in\left[0,100\right]$.

\subsection{Data Setting}

For the synthetic dataset, we simulate the underlying latent functions $f_i(t), i=1,...,10,$ using MGCPs and generate the intensity rates of different units using a sigmoid link function. Then conditioned on this function, we draw training datasets and test datasets \cite{adams2009tractable}. The number of units generated is $N=10$ where we pick the $N$th unit as the testing unit. This experiment is repeated for $Q=1000$ times and we report the average prediction performance of the test dataset for each approach. Figure \ref{fig:fig2} visualizes a sample of intensity rates drawn from the defined MGCP model with four outputs passed through a sigmoid link function along with the simulated events based on the method discussed in \cite{adams2009tractable}. We note that here we only have four outputs for the purpose of illustration while the actual simulation study is done using MGCP model with $N=10$ outputs.

In addition to generating the intensity rates using MGCP and sigmoid link function, we define two parametric functional forms to generate the intensity rates as follows: 

1) A sum of an exponential and a Gaussian bump $\lambda_i(x)=a\exp(-\frac{x}{b})+\exp(-(\frac{x-c}{15})^2)$ where $[a,b,c]^T\sim \mathcal{N}(\bm{\mu}_1,\bm{\Sigma}_1)$ with $\bm{\mu}_1=[3,20,65]^T$ and $\bm{\Sigma}_1=\begin{bmatrix}
5e-1 & 4e-4 & -e-5\\
4e-4 & 2.5e-1 & 3e-7\\
e-5 & 3e-7 & 1
\end{bmatrix}$.

\begin{figure*}[!t]
\centering
\minipage{0.4\textwidth}
  \includegraphics[width=\linewidth]{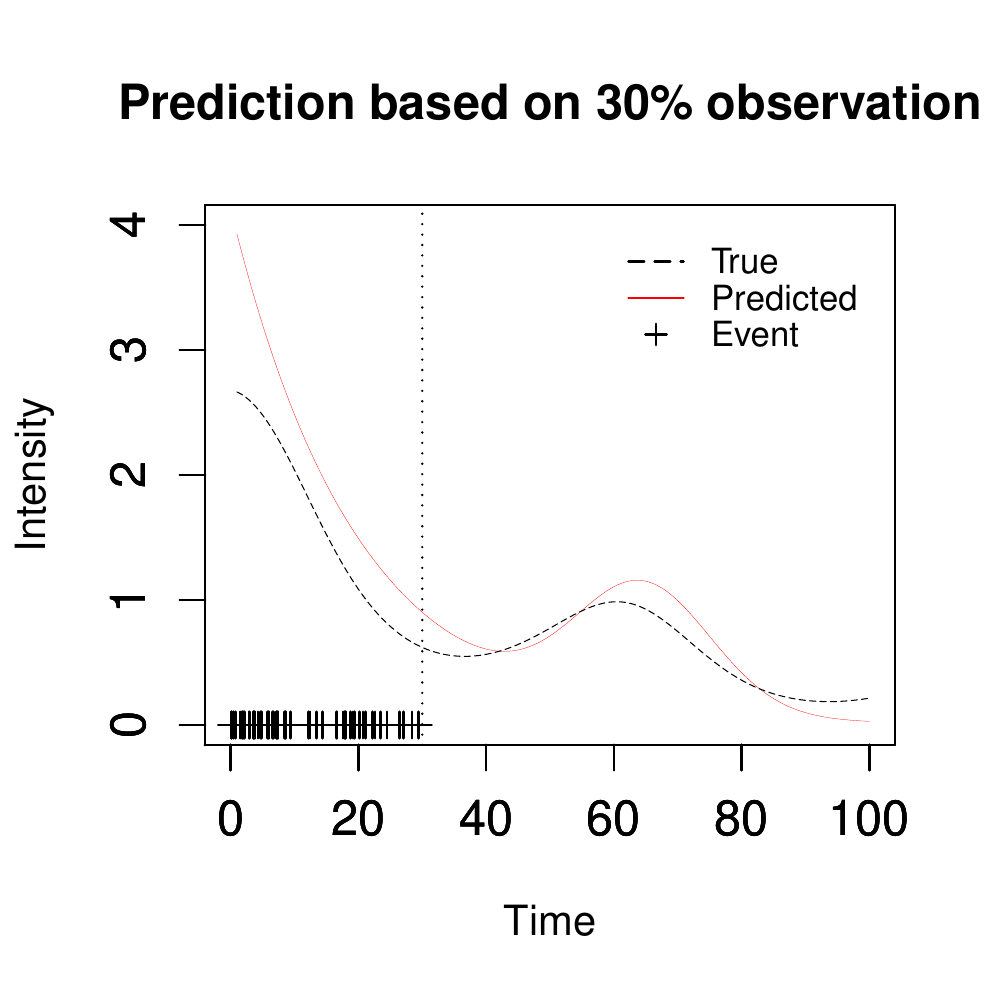}
\endminipage\qquad 
\minipage{0.4\textwidth}%
  \includegraphics[width=\linewidth]{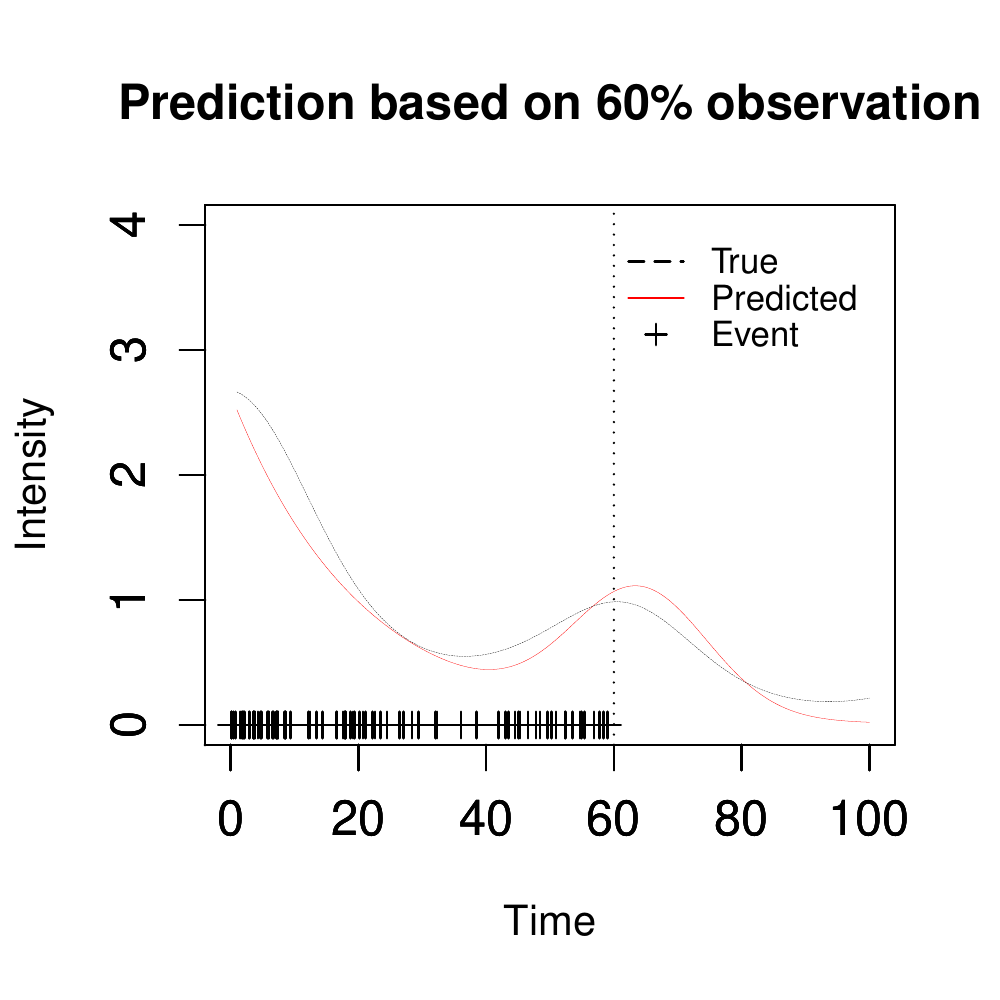}
\endminipage
\caption{Prediction performance for testing unit in different observation percentiles}\label{fig:mot}
\end{figure*}

2) A sinusoid with increasing frequency: $\lambda_i(x)=a' \sin (b' x^2)\exp(-\frac{x}{c'})+1$ where $[a',b',c']^T\sim \mathcal{N}(\bm{\mu}_2,\bm{\Sigma}_2)$ with $\bm{\mu}_2=[2,2e-3,50]^T$ and $\bm{\Sigma}_2=\begin{bmatrix}
1 & -e-7 & 2e-4\\
-e-7 & e-2 & 3e-7\\
e-5 & 3e-7 & 1
\end{bmatrix}$.

We, also, generate the training and testing data conditioned on these functional forms in the simulation study.

\subsection{Results}

We compare different approaches in terms of predictive log-likelihood (LL) and root mean squared (RMS) error between the predicted intensity rate and the true intensity rate of the testing unit $N$. Prediction performance at varying time points $t^\ast$ for the partially observed unit $N$ is reported. The time instant $t^\ast=\alpha \times 100$ is defined as the $\alpha$-observation percentile of the testing unit $N$. The values of $\alpha$ is specified as 30\% and 60\% in the simulations. Figure \ref{fig:mot} illustrates an example of the unit observed up to different percentiles of its life. The intensity rates here are generated using the first parametric functional form explained above. The illustrative example in Figure \ref{fig:mot} demonstrates the behavior of our method. As can be seen from the figure, our joint modeling framework can provide accurate prediction of the true intensity rate for the testing unit $N$. It is mainly because of the flexible convolution structure considered for the MGCP approach that makes sharing of information possible among different units. The unique smoothing kernel $G_{i}$ for each individual allows flexibility in the prediction as it enables each training signal to have its own characteristics. This indeed substantiates the strength of the MGCP. Using the shared latent processes, the model can infer the similarities among all units and predict the intensity rate for the testing unit more accurately by borrowing strength from the training data.

\begin{figure}[!t]
\minipage{0.5\textwidth}
  \includegraphics[width=\linewidth]{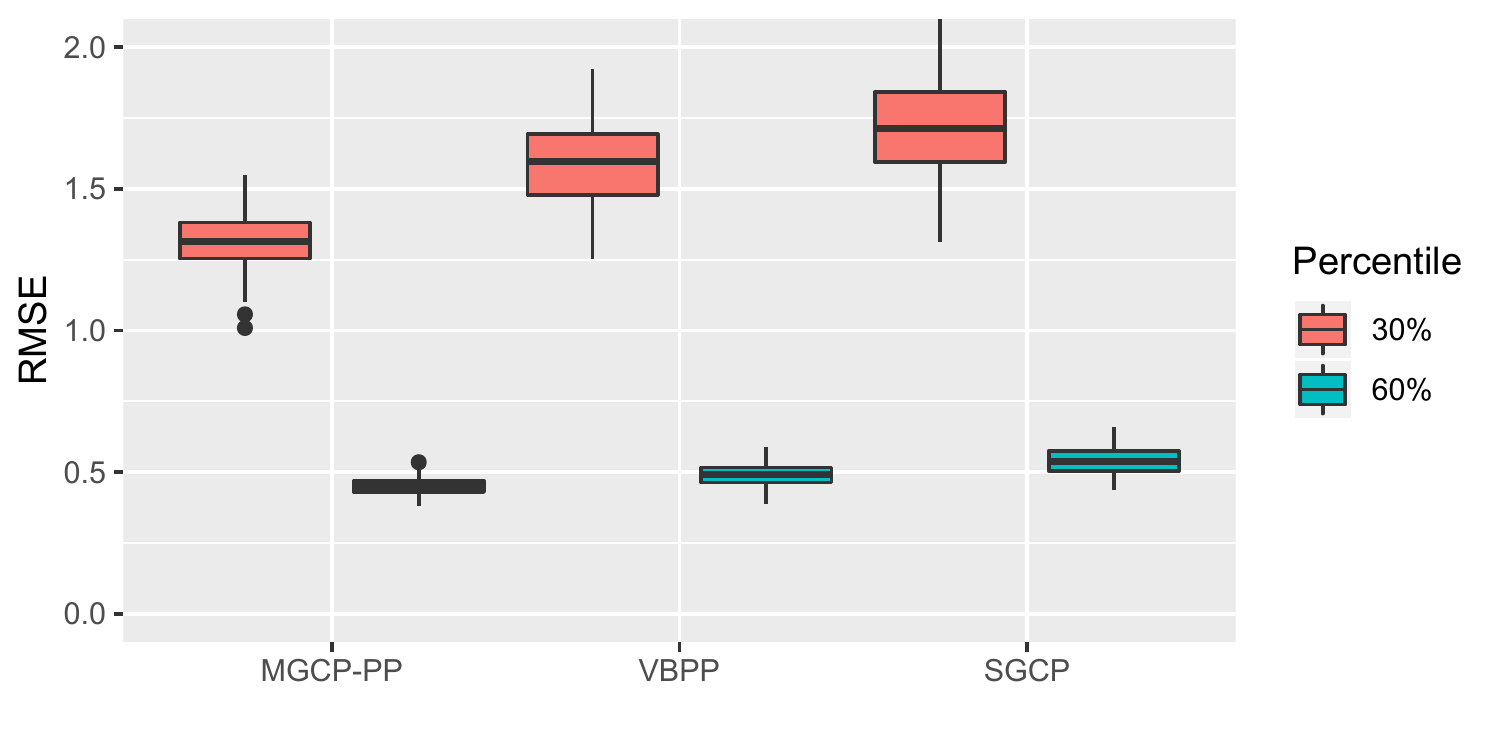}
\endminipage\hfill
\minipage{0.5\textwidth}%
  \includegraphics[width=\linewidth]{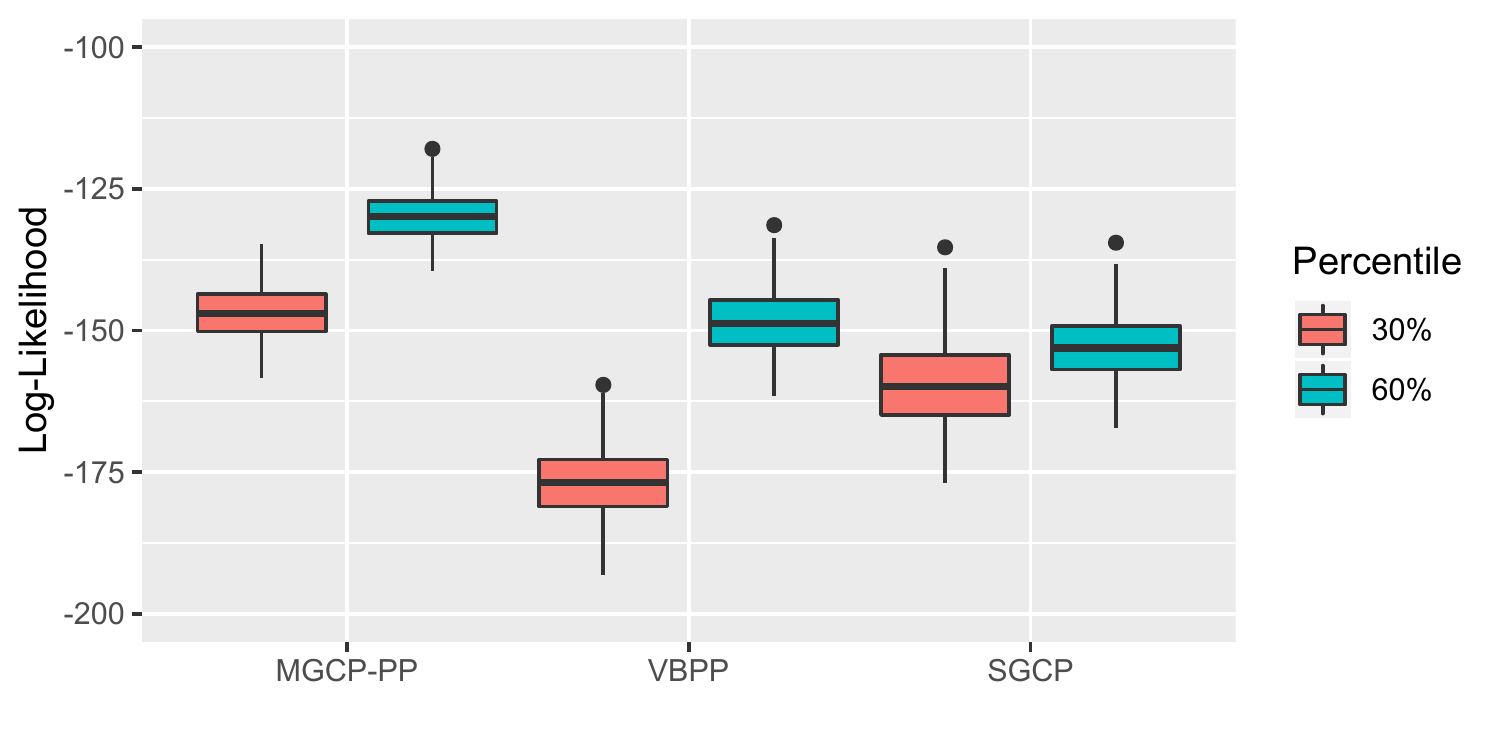}
\endminipage
\caption{Simulation study results with MGCP and the sigmoid link function}\label{fig:sim1}
\end{figure}

\begin{figure}[!t]
\minipage{0.5\textwidth}
  \includegraphics[width=\linewidth]{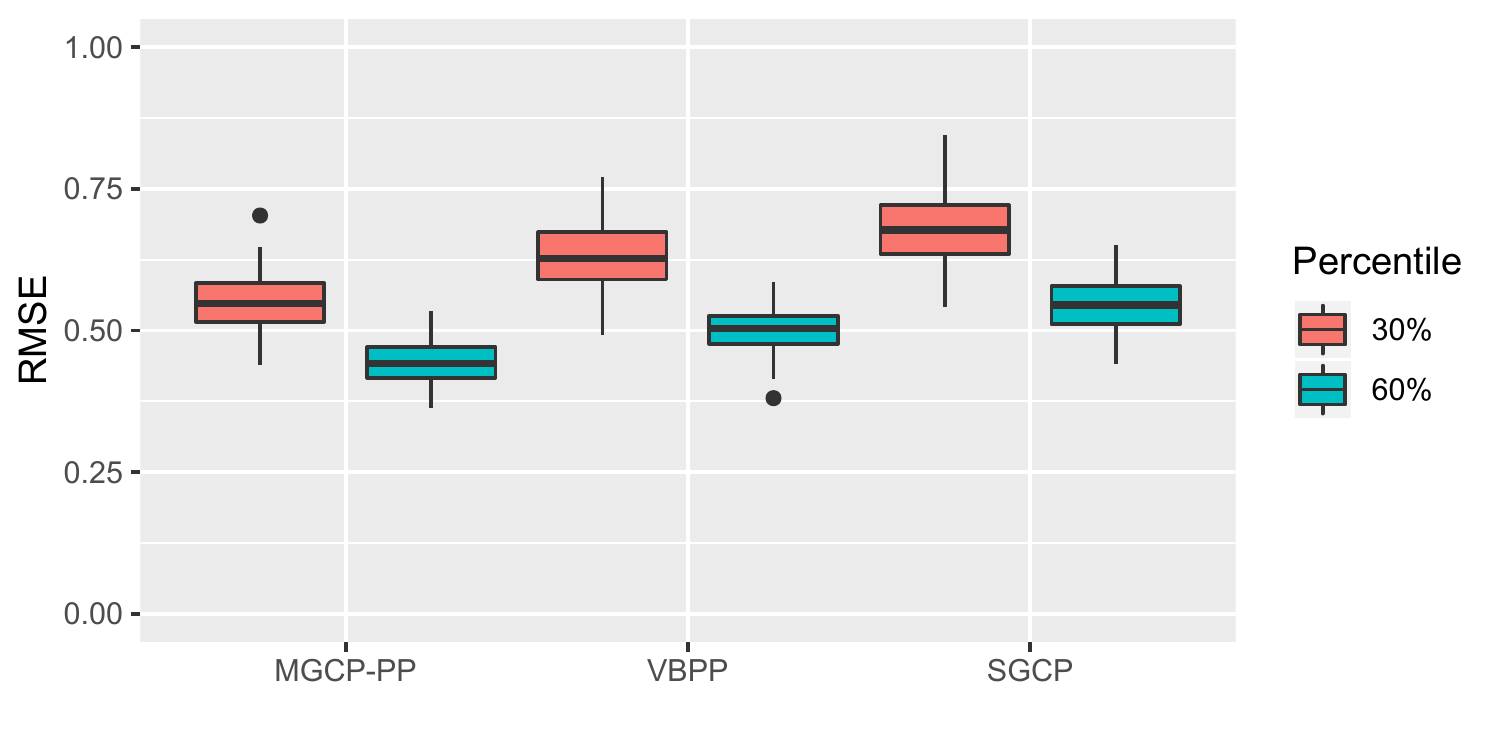}
\endminipage\hfill
\minipage{0.5\textwidth}%
  \includegraphics[width=\linewidth]{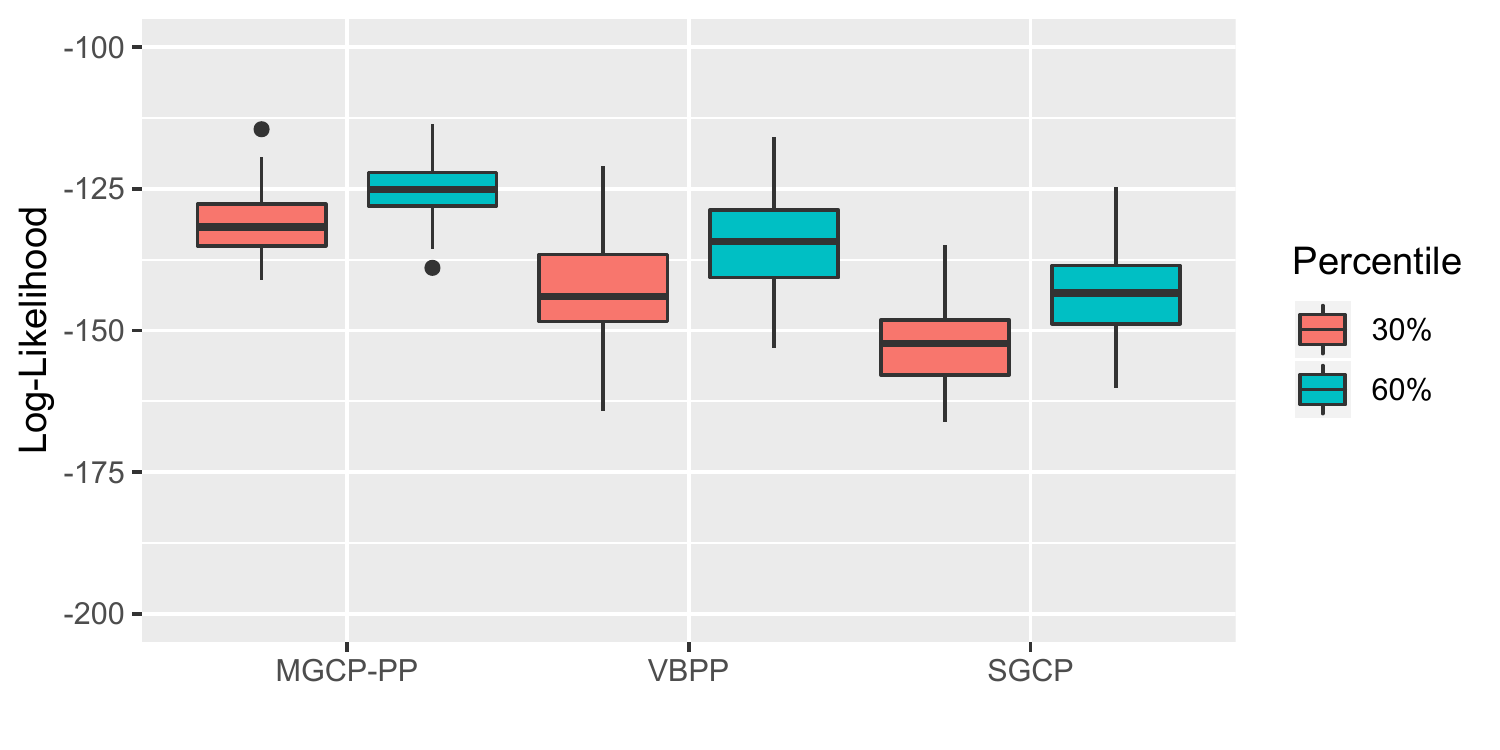}
\endminipage
\caption{Simulation study results with parametric functional form 1}\label{fig:sim2}
\end{figure}

\begin{figure}[!t]
\minipage{0.5\textwidth}
  \includegraphics[width=\linewidth]{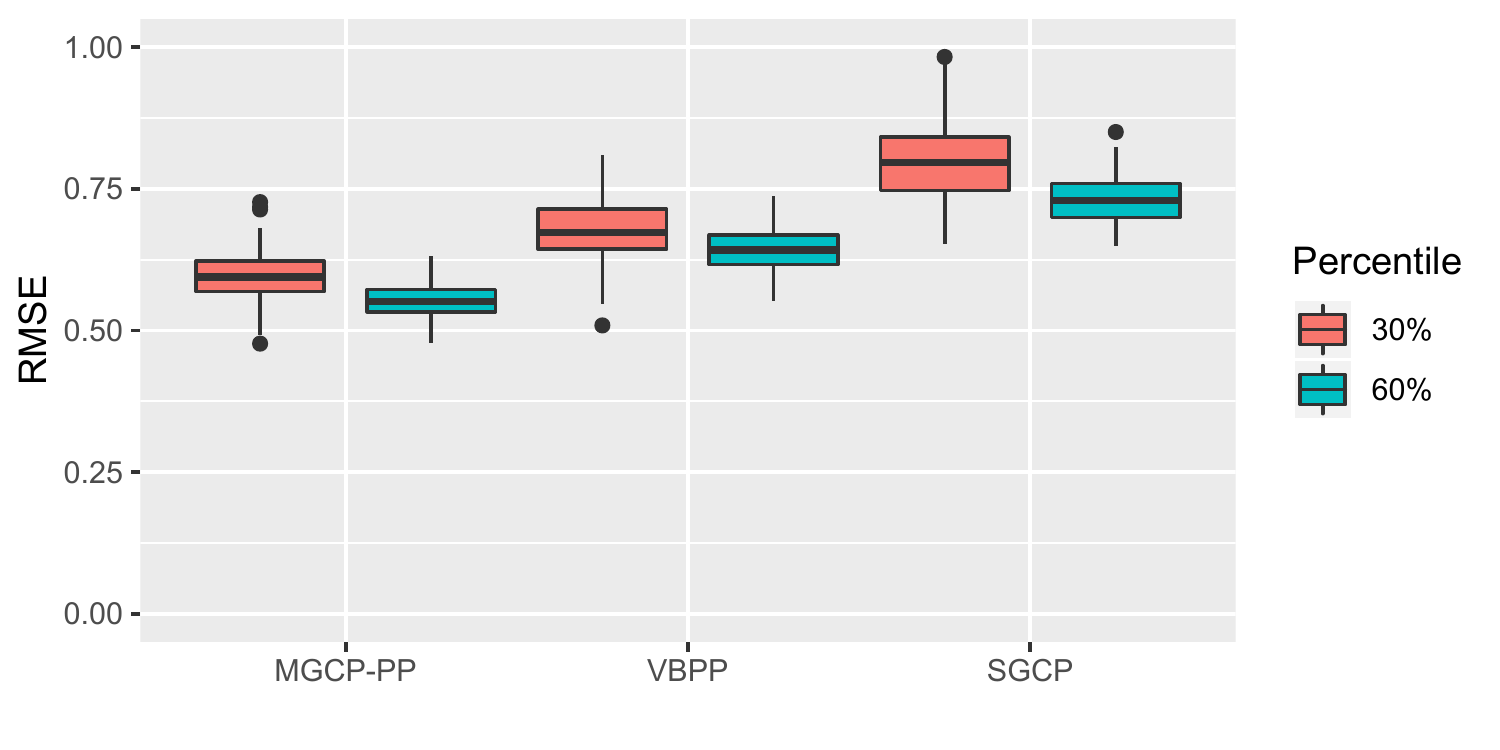}
\endminipage\hfill
\minipage{0.5\textwidth}%
  \includegraphics[width=\linewidth]{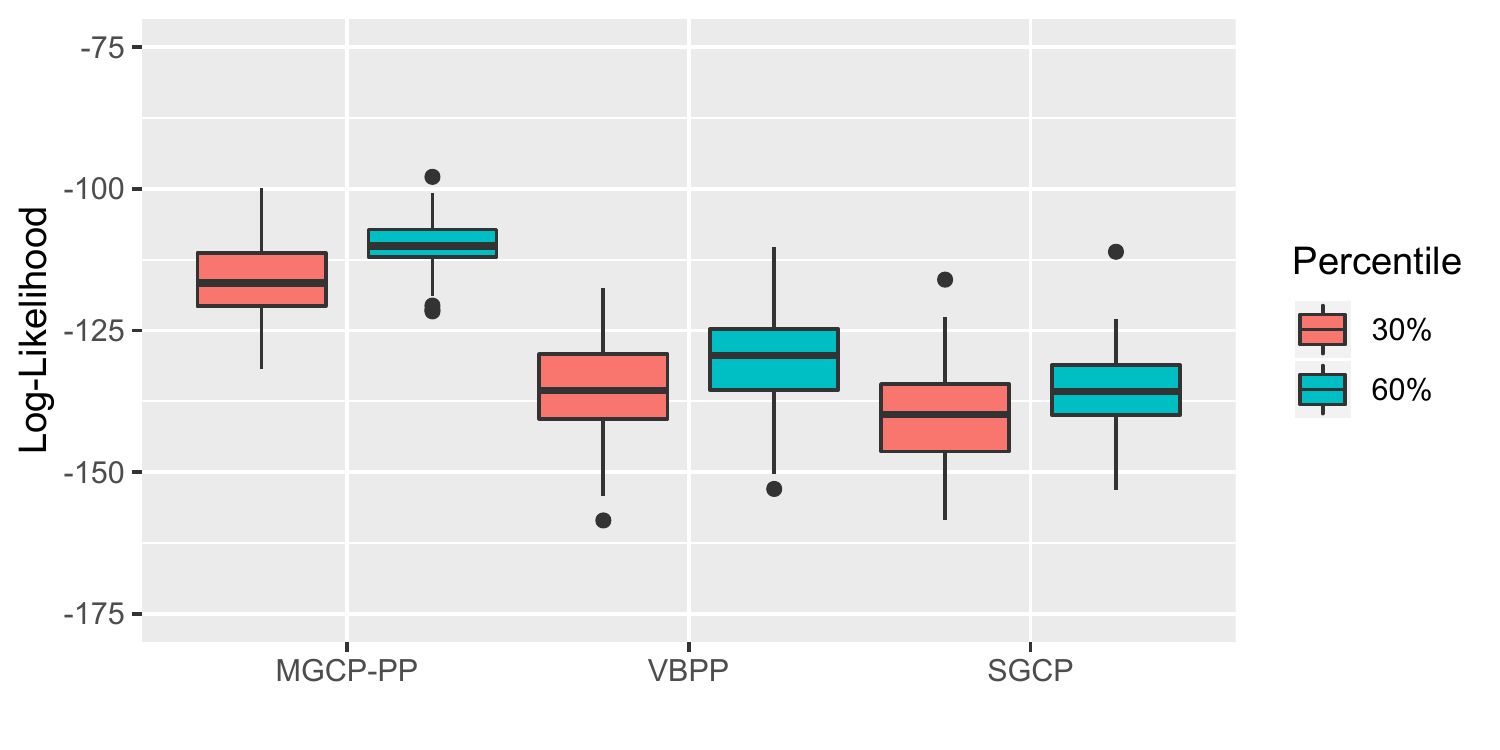}
\endminipage
\caption{Simulation study results with parametric functional form 2}\label{fig:sim3}
\end{figure}

The results in figures \ref{fig:sim1}, \ref{fig:sim2} and \ref{fig:sim3} indicates that our MGCP-PP model clearly outperforms the benchmarked models. Based on the figures we can get some important insights. First, as expected, the prediction errors decrease as the lifetime percentile increase for the testing unit $N$. Thus, the prediction accuracy from the MGCP-PP becomes more accurate as $t^\ast$ increases and more data are collected from the online monitoring unit. Second, we can observe that in the first simulation study, where the data are generated using MGCP and the sigmoid link function, the SGCP approach gives better predictive performance than the VBPP; however, our MGCP-PP model approach always remains superior as it takes advantage of the pool of historical offline units in making inference for the online unit under consideration. The reason that the SGCP approach performs better than the VBPP here can be attributed to the fact that the SGCP uses the same link function and the generative process which results in well-tuned hyper-parameters. Lastly, one striking feature shown in figures \ref{fig:sim1}, \ref{fig:sim2} and \ref{fig:sim3}, is that even with a small number of observations (30\% observation percentile) from the testing unit we are still able to get accurate prediction results. This is crucially important in many applications, specially when observed data are sparse, as it allows early prediction of an event occurrence such as part replacements.

\section{Real-world case study}

In this section, application of the proposed procedure on the real-world data for fleet based event prediction is demonstrated. The data is collected from material handling forklift trucks and events of interest are part replacements captured in real-time. The event occurs across 20 trucks and the number of events for the trucks lies in the range of 6-23. The actual calendar time is adjusted for each unit, i.e. the starting time is made zero for all the units. Figure \ref{fig:casestudy} illustrates the data collected from forklift trucks collected in a teleservice system for warehouse material handling equipment. Please note that the time axis is the lifetime of the forklifts, not the calendar time.

\begin{figure}[ht]
  \includegraphics[width=\linewidth]{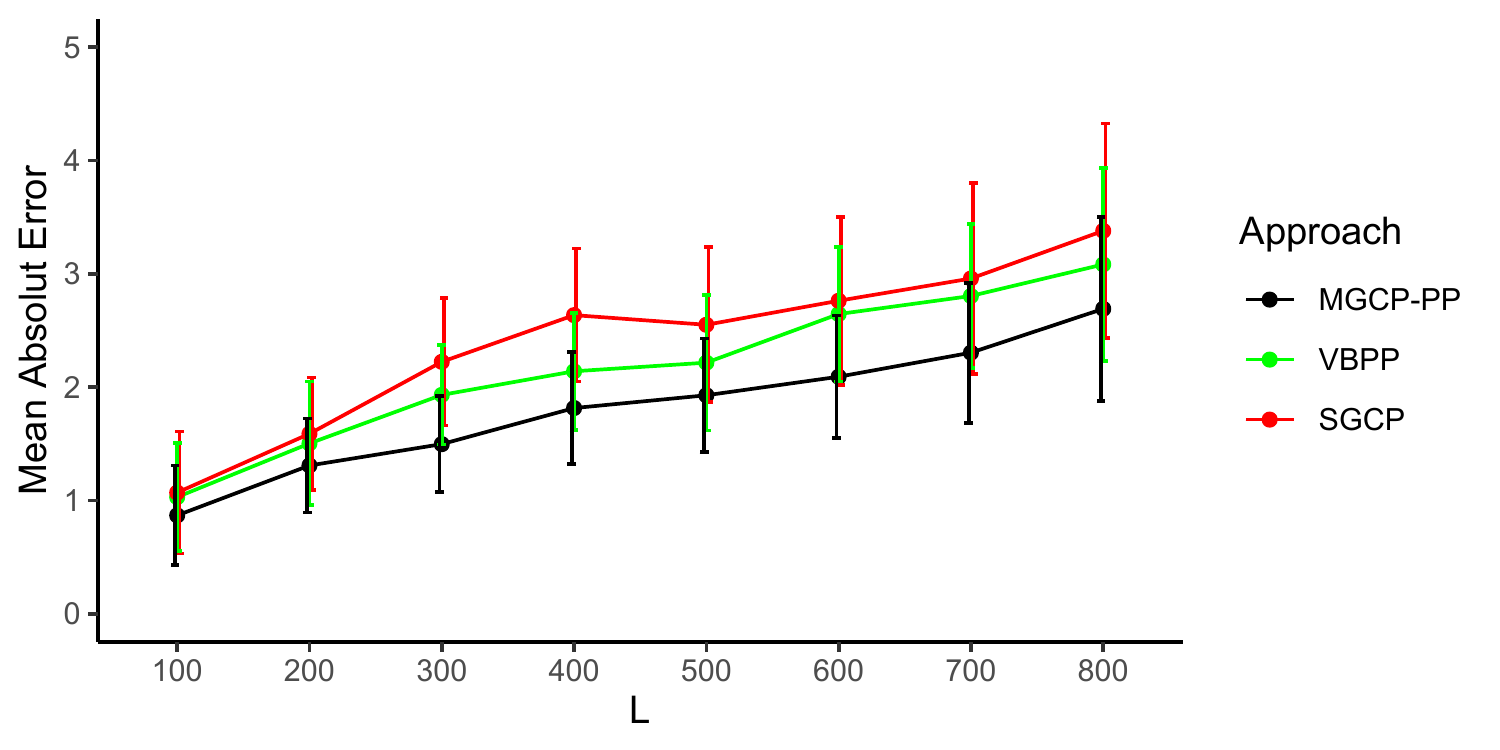}
\caption{MAE of event occurrence in $[t^\ast,t^\ast+L]$.}\label{fig:mae}
\end{figure}

The models MGCP-PP, VBPP, and SGCP are fitted on the case study data.
To perform a comprehensive performance evaluation we use a leave-one-out cross-validation approach. First, we exclude one of the 20 units as the testing unit $N$ and the rest are used as the training units. Prediction for testing unit $N$ is then performed at 50\% lifetime percentile. The whole procedure is repeated 20 times and the predictive performance of different approaches as a function of prediction window $L$ is illustrated in figure \ref{fig:mae} .

Figure \ref{fig:mae} shows the Mean Absolute Error (MAE) of the event occurrence counts in $\left[t^\ast,t^\ast+L\right]$. The MAE of all methods increases monotonically as prediction window length increases. The MGCP-PP approach outperforms the SGCP and VBPP approaches that are based on the univariate Gaussian processes. This indeed highlights the importance of borrowing information from the peer units. Our MGCP-Poisson model that facilitates sharing of information between the testing unit and the training units in the historical datatset clearly extrapolates the intensity rate more accurately which results in better prediction performance.

\section{Conclusion}

In this study, a flexible and efficient non-parametric joint modeling framework for analyzing event data is presented. Specifically, we propose a multivariate Gaussian convolution process modulated Poisson process model that leverages information from all units via a shared latent function. A variational inference framework using inducing variables is further established to jointly estimate parameters from the MGCP-Poisson model accurately. The main advantage of the proposed framework is that it allows accurate individualized prediction for units in the field using the observations from the historical off-line units. The empirical studies highlight the advantageous features of our modeling framework to predict the intensity rates and provide reliable event prediction. 

The model presented in this study can be readily extended to incorporate other observation covariates. Moreover, the convolution structure proposed in this study is flexible. Here, we only shared one latent process across all the units. One can modify this structure by adding more independent latent processes for each unit to improve accuracy in modeling heterogeneity across units. Other structures that share a group of latent processes among selected group of units can be also extended from our model structure. Future work will be aimed towards developing these variational structured Poisson process frameworks.

\ifCLASSOPTIONcompsoc
  \section*{Acknowledgments}
\else
  \section*{Acknowledgment}
\fi

The financial support of this work is provided by The Raymond Corporation and National Science Foundation research grant \#1824761.

\bibliographystyle{IEEEtran}

\bibliography{salmancite}
\end{document}